\documentclass[conference]{IEEEtran}
\IEEEoverridecommandlockouts

\usepackage{cite}
\usepackage{amsmath,amssymb,amsfonts}
\usepackage{algorithmic}
\usepackage{graphicx}
\usepackage{textcomp}
\usepackage{xcolor}
\def\BibTeX{{\rm B\kern-.05em{\sc i\kern-.025em b}\kern-.08em
    T\kern-.1667em\lower.7ex\hbox{E}\kern-.125emX}}
\begin{document}

\title{ProcessGPT: Transforming Business Process Management with Generative Artificial Intelligence
\thanks{Corresponding author: Prof. Amin Beheshti (amin.beheshti@mq.edu.au)\\
Accepted in: 2023 IEEE International Conference on Web Services (ICWS)
}
}

\author{
\IEEEauthorblockN{1\textsuperscript{st} Amin Beheshti*}
\IEEEauthorblockA{\textit{School of Computing} \\
\textit{Macquarie University}\\
Sydney, Australia \\
amin.beheshti@mq.edu.au}
\\
\IEEEauthorblockN{4\textsuperscript{th} Boualem Benatallah}
\IEEEauthorblockA{\textit{School of Computing} \\
\textit{Dublin City University}\\
Dublin, Ireland \\
boualem.benatallah@dcu.ie}
\\
\IEEEauthorblockN{7\textsuperscript{th} Hamid Reza Motahari Nezhad}
\IEEEauthorblockA{\textit{School of Computing} \\
\textit{Macquarie University}\\
Sydney, Australia \\
hamidreza.motaharinezhad@mq.edu.au}
\and
\IEEEauthorblockN{2\textsuperscript{nd}
Jian Yang}
\IEEEauthorblockA{\textit{School of Computing} \\
\textit{Macquarie University}\\
Sydney, Australia \\
jian.yang@mq.edu.au}
\\
\IEEEauthorblockN{5\textsuperscript{th} Fabio Casati}
\IEEEauthorblockA{\textit{School of Computing} \\
\textit{Macquarie University}\\
Sydney, Australia \\
fabio.casati@mq.edu.au}
\\
\IEEEauthorblockN{8\textsuperscript{th} Xuyun Zhang}
\IEEEauthorblockA{\textit{School of Computing} \\
\textit{Macquarie University}\\
Sydney, Australia \\
xuyun.zhang@mq.edu.au}
\and
\IEEEauthorblockN{3\textsuperscript{rd}
Quan Z. Sheng}
\IEEEauthorblockA{\textit{School of Computing} \\
\textit{Macquarie University}\\
Sydney, Australia \\
michael.sheng@mq.edu.au}
\\
\IEEEauthorblockN{6\textsuperscript{th} Schahram Dustdar}
\IEEEauthorblockA{\textit{Distributed Systems Group} \\
\textit{Vienna University of Technology}\\
Vienna, Austria \\
dustdar@dsg.tuwien.ac.at}
\\
\IEEEauthorblockN{9\textsuperscript{th} Shan Xue}
\IEEEauthorblockA{\textit{School of Computing} \\
\textit{Macquarie University}\\
Sydney, Australia \\
emma.xue@mq.edu.au}
}

\maketitle

\begin{abstract}

Generative Pre-trained Transformer (GPT) is a state-of-the-art machine learning model capable of generating human-like text through natural language processing (NLP). GPT~is trained on massive amounts of text data and uses deep learning techniques to learn patterns and relationships within the data, enabling it to generate coherent and contextually appropriate text. This position paper proposes using GPT technology to generate new process models when/if needed. We introduce ProcessGPT as a new technology that has the potential to enhance decision-making in data-centric and knowledge-intensive processes. ProcessGPT can be designed by training a generative pre-trained transformer model on a large dataset of business process data. This model can then be fine-tuned on specific process domains and trained to generate process flows and make decisions based on context and user input. The model can be integrated with NLP and machine learning techniques to provide insights and recommendations for process improvement. Furthermore, the model can automate repetitive tasks and improve process efficiency while enabling knowledge workers to communicate analysis findings, supporting evidence, and make decisions. ProcessGPT can revolutionize business process management (BPM) by offering a powerful tool for process augmentation, automation, and improvement. Finally, we demonstrate how ProcessGPT can be a powerful tool for augmenting data engineers in maintaining data ecosystem processes within large bank organizations. Our scenario highlights the potential of this approach to improve efficiency, reduce costs, and enhance the quality of business operations through the automation of data-centric and knowledge-intensive processes. These results underscore the promise of ProcessGPT as a transformative technology for organizations looking to improve their process~workflows.

\end{abstract}

\begin{IEEEkeywords}
Business Process Management,
Generative AI,
Generative Pre-trained Transformer,
Knowledge-Intensive Processes,
Data-Centric Processes
\end{IEEEkeywords}

\section{Scope and Motivation}

The age of generative Artificial intelligence (AI) is revolutionizing business operations in organizations by augmenting  and automating processes. This necessitates a rethinking of Business Process Management (BPM) as businesses strive to understand the behaviour of their information systems, processes, and services. The proliferation of tools for analyzing process executions, system interactions, and system dependencies is evidence of this priority in medium and large enterprises~\cite{ProcessAnalytics,van2013business}.
With the advent of generative AI, businesses can use AI augmentation to enhance their processes in ways previously unimaginable~\cite{barukh2021cognitive}. For instance, content generation using generative AI is changing many processes, such as customer service, graphic design, and journalism, while projects like Github Copilot\footnote{https://github.com/features/copilot} are changing the software development industry.

The use of generative AI technologies such as generative pre-trained transformer (GPT)~\cite{radford2019language} can help maintain process models in process silos using generative AI-enabled approaches, especially in ad-hoc processes where it would be challenging for knowledge workers to choose the best next step. In this context, the focus should shift towards understanding and analyzing business process-related data captured in various information systems and services that support processes while rethinking the traditional BPM methodologies.

\subsection{Generative Pre-trained Transformer}

Generative Pre-trained Transformer (GPT) is a state-of-the-art language model that utilizes deep neural networks to generate human-like text. The model was first introduced by OpenAI\footnote{https://openai.com/} in 2018 and has since been updated with larger and more complex versions. GPT's success in generating coherent and high-quality text has led to the development of other generative models that employ similar techniques.
Examples include:
Text Generation:
ChatGPT\footnote{https://openai.com/blog/chatgpt}  developed by OpenAI;
Image Generation:
DALL-E\footnote{https://openai.com/product/dall-e-2} developed by OpenAI;
Speech Generation:
VALL-E\footnote{https://valle-demo.github.io/} developed by Microsoft; and
Code Generation:
GitHub Copilot\footnote{https://github.com/features/copilot} developed by GitHub\footnote{https://github.com/} and OpenAI.

The potential of generative AI extends beyond just generating new data, as it has the capability to revolutionise the processes in organizations and personal life by offering process augmentation, automation, and improvement.
Generative AI has numerous applications, including the automation of content creation tasks such as writing articles, creating social media posts, and generating product descriptions. Additionally, it can be used to identify patterns and anomalies in financial transactions, help to detect and prevent fraudulent activities, or automatically inspect and evaluate the quality of products, reducing the need for manual inspection. Another benefit of generative AI is that it can automate customer service tasks, such as answering common queries and handling routine tasks, allowing human agents to focus on more complex issues.

\subsection{Business Process Management}

Business Process Management (BPM) refers to the discipline of managing, modelling, analyzing, and improving business processes within an organization~\cite{ProcessAnalytics,dumas2013fundamentals}. Business processes are central to the operation of public and private enterprises. They involve activities that transform inputs into outputs, delivering value to customers and stakeholders. The success of an enterprise depends on how efficiently and effectively its processes are executed. For this reason, business process analytics has always been a key endeavour for companies~\cite{ProcessAnalytics,beheshti2022social}.
Early efforts to address this goal started with process automation, where workflow and other middleware technologies were used to reduce human involvement through better systems integration and automated execution of business logic. The total or partial automation of the process creates an unprecedented opportunity to gain visibility on process executions in the enterprise.

In addition to the techniques, frameworks, and methodologies mentioned for business process management, there is an increasing need to support knowledge workers in ad-hoc processes~\cite{beheshti2022ai}. These workers require technologies such as GPT to help them offer new processes by maintaining process models in process silos using generative AI-enabled approaches. This is particularly important in scenarios where it is challenging for the knowledge worker to choose the best next step. With the help of these technologies, the focus of process thinking has expanded to include the creation and maintenance of process models that can adapt to changing business needs and support decision-making in real time. These advancements have further increased the importance of analyzing business process-related data to discover helpful information and achieve business objectives.

In our previous work~\cite{beheshti2022ai}, we introduced the design of AI-enabled business processes.
A recent approach, ProcessGAN~\cite{van2023processgan}, has been proposed to employ generative adversarial networks (GANs) for business process improvement (BPI). This approach assists process designers in creating innovative BPI concepts.
Another approach, proposed by Mustansir et al.~\cite{mustansir2022towards}, used an NLP-based approach to extract redesign suggestions for automatic business process redesign.

\subsection{Data-centric Processes}

Analyzing process-related data is crucial in data-centric processes, which are gaining importance in medium and large enterprises. This data analysis helps enterprises to identify business needs, determine solutions to problems, and make informed decisions. Process data analytics has seen significant advancements in recent years, with research focusing on linking the insights gained from process data analytics to the process analysis phase. By analyzing data from various sources such as enterprise systems, sensor data, and social media, companies can gain insights into their processes and improve them for better performance. Examples of data-centric processes include supply chain management, fraud detection, and customer experience optimization. These practices and technologies provide process analytics from querying to analyzing process data, thus covering a broad spectrum of business process paradigms.

\subsection{Knowledge-Intensive Processes}

Knowledge-intensive processes encompass operations that critically depend on the expertise of subject-matter experts, whose cognitive capabilities have been shaped by the acquisition and accumulation of knowledge and experience within their biological neural networks~\cite{zhao2022deep}.
This source of knowledge can be used as the core of AI-enabled Processes, for example, to facilitate understanding the intentions of end-users better or providing a comprehensive method to retrieve useful information from the enterprise platform.
In our previous work~\cite{KB40}, we proposed the use of crowdsourcing services for mimicking the knowledge of domain experts in knowledge-intensive processes,
and use this knowledge to build a new type of Knowledge Bases (namely Knowledge Base 4.0) that can facilitate the auto labelling of the data to be used in learning algorithms. The goal is to facilitate knowledge sharing from domain experts using rule-based crowdsourcing services.

In this paper, we take a step forward in introducing the concept of ProcessGPT. This model is specifically designed to cater to data-centric and knowledge-intensive processes. It achieves this by training a generative pre-trained transformer model on a vast amount of business process data and models.
This model can then be fine-tuned on specific process domains and trained to generate process flows and make decisions based on context and user input. The model can also be integrated with natural language processing and machine learning techniques to provide insights and recommendations for process improvement. Furthermore, the model can be used to automate repetitive tasks and improve process efficiency, while also enabling knowledge workers to communicate analysis findings, support evidence, and make decisions. ProcessGPT has the potential to revolutionize business process management by offering a powerful tool for process automation, improvement, and decision-making.

\begin{figure*}[t]
	\begin{center}
	\includegraphics[width=1.0\linewidth]{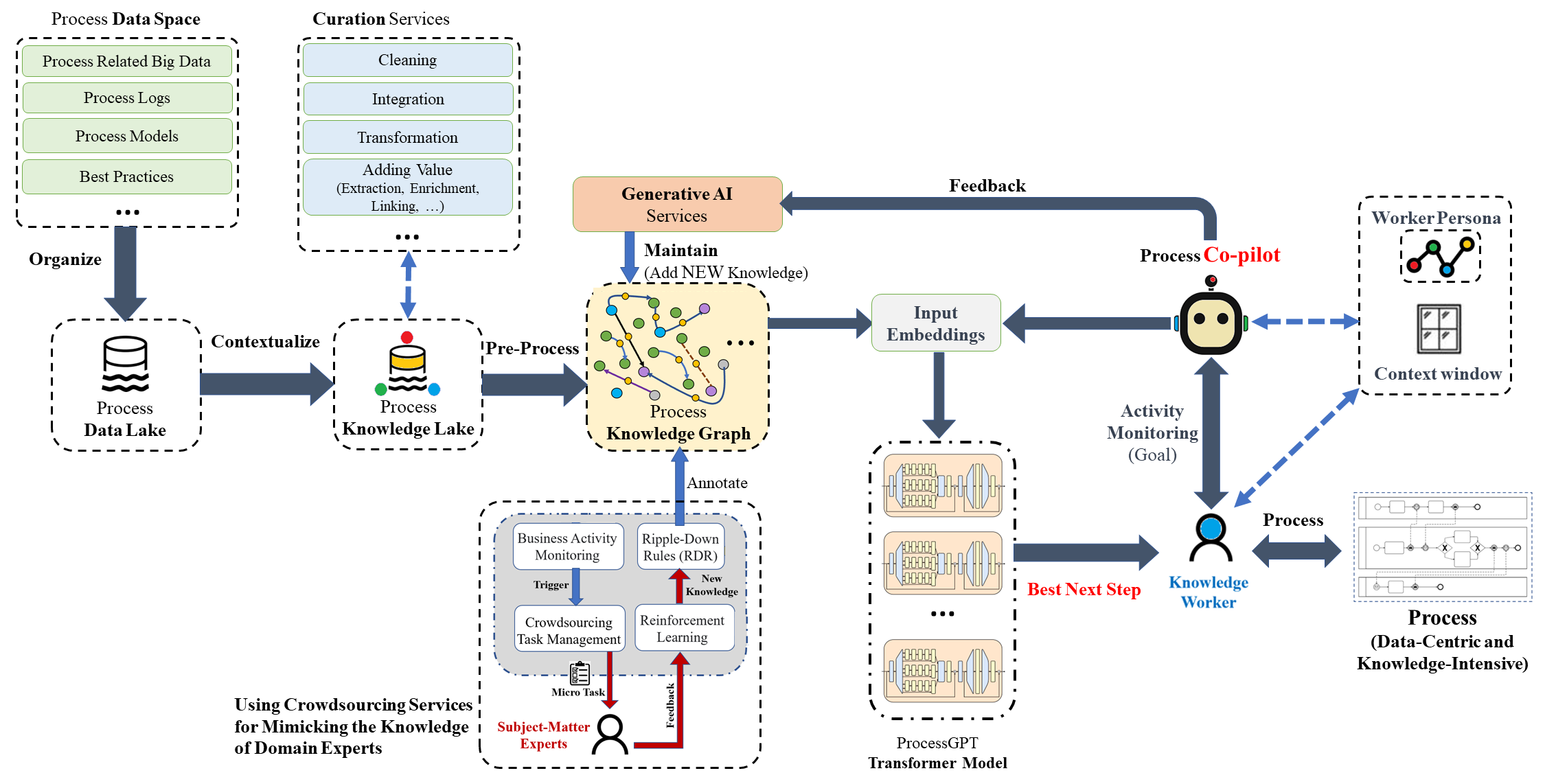}
	\end{center}
	\caption{Proposed Architecture for ProcessGPT.}
	\label{ProcessGPT}
\end{figure*}

\section{ProcessGPT: From Augmentation to Automation}

The proposed architecture for ProcessGPT is depicted in Figure~\ref{ProcessGPT}, which comprises several essential components such as the Process Data space, Process Data Lake, Process Knowledge Lake, Process Knowledge, Process Knowledge Graph, Transformer Model, and the Process Co-pilot.

The \textbf{Process Data Space} layer refers to all process-related data that is generated across multiple data islands, including process logs, process model silos, best practices, and process-related data that is generated across various data islands such as open, private, and social data. This data exhibits typical properties of big data, including wide physical distribution, diversity of formats, nonstandard data models, and independently managed and heterogeneous semantics.
To manage this large and diverse process-related data, the \textbf{Process Data Lake} layer is proposed. This layer leverages our previous work, Data Lake as a Service~\cite{CoreDB}, to organize, index, and query the big process data and metadata. The Process Data Lake layer will facilitate the effective organization of the process-related data and metadata to enable efficient retrieval, analysis, and sharing of knowledge across various data islands.


After storing the raw data in the Process Data Lake, the next step is to transform this data into contextualized data and knowledge. This transformation is achieved using the Knowledge Lake as a Service approach, which has been developed in previous work~\cite{CoreKG}.
To facilitate this transformation, the proposed approach involves the creation of a \textbf{Process Knowledge Lake} layer, which utilizes curation services to clean, integrate, and transform the raw data. Additionally, value is added to this data through the extraction, enrichment, and linking of information items.
The contextualized data is pre-processed in the Knowledge Lake to form a large graph known as the \textbf{Process Knowledge Graph}. This graph is a representation of the relationships between the various entities in the process-related data, providing a more complete and holistic view of the process knowledge. The Process Knowledge Graph serves as a foundation for subsequent analyses, enabling effective decision-making and process improvement.

The Process Knowledge Graph is a large, semantically rich graph database that represents information and knowledge through a set of nodes and edges. The nodes in the graph represent entities or concepts, while the edges represent the relationships between these entities. The Process Knowledge Graph serves as a repository for contextualized process data, knowledge, tasks, and activities.
The Knowledge Graph is a powerful tool for representing and organizing knowledge in a structured and interconnected manner. It enables the system to link related concepts and entities together, facilitating reasoning and inference across different domains. This interconnectedness between the various entities in the knowledge graph enables the system to reason about relationships and dependencies between different entities, enabling more effective decision-making and process improvement. The Knowledge Graph represents a significant advancement in process management and provides a foundation for developing intelligent and adaptive process management systems.

The Process Knowledge Graph does not include the best practices and knowledge of domain experts. To address this gap, our previous work~\cite{KB40} has been leveraged to mimic the knowledge of subject-matter experts using crowdsourcing services. This new knowledge is used to annotate the nodes and relationships in the Process Knowledge Graph, enhancing its completeness and accuracy.
At this stage, the knowledge graph serves as a comprehensive source of data for the AI engine, specifically the \textbf{ProcessGPT Transformer Model}.
The Transformer is a novel type of neural architecture that utilizes the attention mechanism to encode input data into robust features.
ProcessGPT will use the Transformer-in-Transformer (TNT)~\cite{han2021transformer} model that highlights the significance of attention within local patches for developing high-performing transformers.
This model utilizes a similar approach to that used in generative pre-trained transformers (GPT), such as chatGPT and Github Co-pilot.

For example, chatGPT aims to suggest the best next keyword in the generated text.
Github Co-pilot aims to suggest the best next line of code in the generated program.
However, the \emph{goal of the ProcessGPT model} is to suggest the best next step in the current process pipeline, leveraging the rich contextual information captured in the Process Knowledge Graph.
In particular, integrating expert knowledge and the ProcessGPT Transformer Model enables more effective decision-making and process optimization, enhancing the overall performance and efficiency of the process management system.

The architecture includes an essential component known as \textbf{Process Co-pilot}, which utilizes the latest advancements in Business Activity Monitoring~\cite{tax2017predictive}. The Process Co-pilot continuously monitors the activities in the current process instance, while simultaneously having access to the Knowledge Worker's Persona and the Context Window, i.e., the number of previous tokens that are considered when generating the next token in the output sequence. It understands the Knowledge Worker's objectives and requirements and sends a query to the \textbf{Input Embeddings} component, which is also fed by the Process Knowledge Graph.
The Input Embeddings component converts the input data into a series of vectors that can be processed by the ProcessGPT transformer model. This model provides the best next step in the processing pipeline, which is suggested to the Knowledge Worker while dealing with the case. This scenario is similar to Github Co-pilot, which suggests the best next code to a programmer in real-time while they are writing the program.
The integration of the Process Co-pilot, Knowledge Worker Persona, Context Window, Input Embeddings component, and ProcessGPT transformer model facilitates more efficient and effective process management, enhancing the performance of the system. The utilization of state-of-the-art technologies and techniques ensures that the system can adapt to the evolving needs and requirements of the process management domain.

The Feedback loop is a crucial step in the architecture. In cases where the Knowledge Worker is not satisfied with the suggested best next step, the Process Co-pilot detects this and sends the feedback (e.g., useful or not useful) to the \textbf{Generative AI Services} component. This provides the generative AI component with access to a rich source of information and focuses on maintaining the Process Knowledge Graph by identifying and correcting errors and inconsistencies in the data and expanding the graph by identifying new nodes and relationships.
This can be achieved by training generative AI models to recognize patterns and trends in the graph data and generate new nodes or relationships to fill gaps or correct inaccuracies. The use of generative AI models enables the system to continually improve and enhance the Process Knowledge Graph, ensuring that it remains up-to-date and accurate. This, in turn, enables the system to provide better suggestions for the best next step in the processing pipeline, leading to improved performance and increased efficiency.

Various methods can be employed to maintain the Process Knowledge Graph, including graph neural networks (GNN)~\cite{zhou2020graph}, generative adversarial networks (GANs)~\cite{creswell2018generative}, and variational autoencoders (VAEs)~\cite{simonovsky2018graphvae}. For instance, if the generative AI model detects an erroneous relationship between two nodes or a missing data point, it can rectify the error by generating a new node or relationship. Furthermore, generative AI can facilitate the expansion of the Knowledge Graph by identifying new nodes and relationships to be added to the graph. In the case of detecting a new concept or activity related to an existing node in the graph, the generative AI model can create a new node and/or relationship to incorporate the new information.
In order to maintain a high-quality approach, the generative AI Service can benefit from an internal feedback loop that shares samples of the new content generated by the model. This feedback loop can help to identify and correct any errors or inconsistencies in the generated content and ensure that the model continues to produce high-quality output. By incorporating this feedback loop, the generative AI Service can continuously improve its performance and better support the knowledge worker in their tasks.

\subsection{Process Augmentation}

Generative AI has the potential to augment knowledge workers in knowledge-intensive processes and amplify their knowledge and skills to be more productive. One such example is the GitHub Copilot project, which uses generative AI to help software engineers write code more efficiently.
GitHub Copilot does this by analyzing the code that a programmer is working on and then suggesting the next line of code based on what it has learned from other code examples (developed by other developers). GitHub Copilot is based on the OpenAI Codex, a generative AI language model that has been trained on a large dataset of code.
While GitHub Copilot is a great example of how generative AI can be used to augment knowledge workers in the software engineering domain, we believe that there is potential for similar technology in other data-centric and knowledge-intensive processes.

Our vision is to utilize ProcessGPT for the creation of domain-specific Co-pilots catering to knowledge workers operating in diverse domains such as health, banking, and education.
Accordingly, this paper proposes the utilization of Generative Pre-trained Transformer (GPT) technology to develop new process models in data-centric and knowledge-intensive processes.
ProcessGPT can be a revolutionary technology for process management by offering powerful support for process automation, improvement, and decision-making.
For example, in the healthcare domain, ProcessGPT can aid doctors in diagnosing patients more efficiently by analyzing patient data and suggesting the most likely diagnosis and treatment options. Similarly, in the banking domain, ProcessGPT can be used to suggest new process/data models to identify potential cases of fraud or money laundering by analyzing transaction data and suggesting suspicious transactions for further investigation. In education, as another example, ProcessGPT can assist teachers in understanding and analyzing students' creativity pattenrs and offer new methods to help students who need help.

\subsection{Process Automation}

The automation of business processes has become increasingly important in today's fast-paced and competitive environment. Companies are looking for ways to improve their efficiency, reduce costs, and increase productivity. One promising approach is to use generative AI technology to automate business processes.
ProcessGPT is designed to be a flexible and customizable tool that can be trained on large business process data and model datasets. The model can be fine-tuned on specific process domains, trained to generate process flows, and offer to automate structured and case-based business processes.
By automating repetitive tasks and providing insights for process improvement, ProcessGPT can amplify the knowledge and skills of knowledge workers and enable them to focus on higher-level tasks. Furthermore, ProcessGPT can reduce errors and improve the quality of decision-making, leading to better outcomes for businesses and their customers.

\section{ProcessGPT for Maintaining Data Ecosystem Processes: An Augmentation Scenario}

Consider a Bank as a large manufacturing company that relies heavily on its data ecosystem for its operations. The company has implemented various processes for data management, data analysis, and reporting. The processes involve several steps and can be quite complex. The company has hired a team of analysts to maintain these processes and ensure that they are running smoothly.
One of the challenges that the analysts face is that the processes are constantly evolving, and they need to update the process models to reflect these changes. This can be a time-consuming and tedious task, as it involves analyzing large amounts of data and identifying the changes that need to be made.

To address this challenge, the bank has decided to implement ProcessGPT. The company trains a generative pre-trained transformer model on a large dataset of business process data, including data from its own ecosystem and the subject-matter experts' knowledge (captured in the Knowledge Base 4.0). The model learns to identify patterns and relationships in the data, which can be used to generate new process models.
%
%
By using ProcessGPT, the bank is able to streamline its process maintenance tasks and reduce the time and effort required to update process models. This enables the analysts to focus on more strategic tasks and ensures that the company's data ecosystem remains optimized for its operations.

\section{ProcessGPT for Automating Exam Marking: An Automation Scenario}

In the education sector, evaluating student performance and providing formative feedback is critical for improving students’ learning and achievements. However, grading exams and assignments can be time-consuming, particularly in large-scale online learning environments. One potential solution to reduce instructors' workload is the use of ProcessGPT to automate exam marking.
In our recent work~\cite{AminPatent}, we leveraged generative AI to automate assessment marking in the education domain.
The benefits of automating exam marking using ProcessGPT are manifold. Firstly, it can save time and resources for educators, allowing them to focus on other critical tasks such as analyzing student performance data and developing personalized learning plans. Secondly, it can eliminate human bias and ensure that all students are graded fairly and accurately. Thirdly, it can provide immediate feedback to students, allowing them to learn from their mistakes and improve their performance in future exams.

With the growing use of technology in education, automating exam marking using ProcessGPT is a logical step towards enhancing the learning process. Moreover, the use of ProcessGPT for exam marking has the potential to significantly improve the quality and reliability of assessments in the education sector. By training ProcessGPT algorithms on a large dataset of previously graded exams, the system can learn how to identify and evaluate specific patterns in the answers, learn from instructors' experience, and accurately and consistently grade exams with much faster turnaround times than human graders.
Moreover, the use of ProcessGPT for exam marking can also detect plagiarism. By analyzing and comparing the answers provided by students in the same exam and previous exams, ProcessGPT algorithms can identify similarities that may indicate plagiarism. This is especially useful for large classes where it may be difficult for human graders to detect such cases. Furthermore, ProcessGPT can also detect subtle differences in writing styles, making it more difficult for students to evade detection. This will ensure academic integrity and discourage cheating, providing a fairer and more transparent evaluation of student performance.

\section{ProcessGPT for Workflows: A Hybrid Scenario for Blending Automation and Augmentation}

In the age of generative AI, the landscape of workflows is undergoing a significant transformation. The emergence of advanced AI models, such as ProcessGPT, can revolutionise how we perceive, design, and utilize workflows.
ProcessGPT paves the way for a transformative approach that seamlessly integrates automation and augmentation, redefining the workflows of the future. This in turn has the potential to revolutionize how tasks are accomplished, offering a hybrid model that combines the efficiency of automation with the expertise of human knowledge workers.
Let us explore a motivating scenario in a police investigation where ProcessGPT showcases its capabilities in supporting a hybrid mode.
Law enforcement agencies operate within a complex web of processes, from evidence collection and suspect identification to case management and legal proceedings. ProcessGPT has the potential to offer a transformative approach to these workflows, enabling automation and, at the same time, acting as a knowledgeable copilot for law enforcement professionals.

\emph{Automation}. Law enforcement workflows encompass several crucial stages: data aggregation, evidence gathering, data analysis, suspect profiling, and decision-making. In the initial stages, investigators spend significant time sifting through vast amounts of data from diverse sources, such as crime databases, surveillance footage, and witness statements. ProcessGPT can automate this data aggregation, rapidly organizing and contextualizing the information, saving investigators valuable time and effort.
For example, by automating the data aggregation process, ProcessGPT significantly reduces the time and effort required for investigators to compile evidence. Investigators can focus on analyzing the extracted information and connecting the dots, rather than spending hours locating and gathering data. This automation expedites the investigation process and enhances the accuracy and comprehensiveness of evidence collection. Furthermore, ProcessGPT's ability to contextualize the gathered information adds value to workflow automation. It can analyze the relationships between different pieces of evidence, identify patterns, and highlight potential leads for further investigation.

\emph{Augmentation}. By leveraging its comprehensive knowledge base and reasoning abilities, ProcessGPT can generate insightful questions and alternative investigative strategies and provide contextual guidance to investigators. This augmentation aspect helps investigators uncover hidden connections, identify patterns, and consider perspectives that might have been overlooked, enhancing their decision-making capabilities.
Additionally, ProcessGPT's ability to learn from successful investigation patterns, legal precedents, and evolving crime trends empowers it to improve its analytical skills continuously. This evolution ensures that ProcessGPT remains a reliable resource, guiding investigators through complex cases by suggesting relevant leads, recommending potential courses of action, and keeping them updated with the latest developments in the field.

\section{Summary and Discussion}

In this position paper, we proposed using Generative Pre-trained Transformer (GPT) technology to generate new processes to facilitate decision-making in data-centric and knowledge-intensive processes. ProcessGPT can be designed by training a generative pre-trained transformer model on a large dataset of business process-related data and models, and then fine-tuning it on specific process domains to generate process flows and make decisions based on context and user input. Through the integration of natural language processing and machine learning techniques, the model can provide insights and recommendations for process improvement. Additionally, the model can automate repetitive tasks, improve process efficiency, and enable knowledge workers to communicate analysis findings and make informed decisions.

The application of ProcessGPT in business process management can bring a transformative change.
We are exploring the full potential of ProcessGPT in different domains, such as healthcare, finance, and manufacturing. Additionally, a deep dive into the ethical and legal implications of ProcessGPT will be explored. Another line of research focuses on the scalability and sustainability of ProcessGPT to ensure its long-term viability as a transformative technology for organizations looking to improve their process workflows.
In the following, we discuss a few future directions and ongoing works.

\emph{Understanding Conversations}.
ProcessGPT can benefit from improved natural language understanding capabilities, allowing it to more accurately comprehend and interpret complex instructions, queries, and context. Advancements in this area would enable more precise and context-aware automation and augmentation.
As an ongoing work, we are exploring various techniques such as semantic parsing, entity recognition, and coreference resolution~\cite{CDCR} to extract deeper meaning from conversations. Additionally, advancements in natural language understanding can be coupled with contextual reasoning and world knowledge integration to enhance ProcessGPT's comprehension of specific domains.

\emph{Domain-Specific Knowledge}.
ProcessGPT can be further developed to incorporate domain-specific knowledge and expertise, making it more effective in supporting specialized workflows. By integrating specific industry knowledge, regulations, and best practices, ProcessGPT can provide tailored guidance and automation in various sectors such as healthcare, education, finance, and law enforcement.

\emph{Advanced Decision Support}.
ProcessGPT can evolve to offer advanced decision support by incorporating probabilistic reasoning, risk analysis, and predictive modelling. This would enable it to assist knowledge workers in complex decision-making processes, providing insights and recommendations based on sophisticated analysis of data and patterns.

\emph{Multi-Modal Capabilities}.
Expanding the capabilities of ProcessGPT to handle multi-modal inputs, such as text, images, videos, and audio, would open up new possibilities for automation and augmentation. This would allow ProcessGPT to analyze and interpret information from diverse sources, enriching its understanding and enhancing its ability to support complex workflows.

\emph{Collaborative Workflows}.
Future developments in ProcessGPT could focus on facilitating collaborative workflows, enabling seamless cooperation and interaction between human users and the AI system. This would involve real-time collaboration, shared decision-making, and synchronized task management, fostering a productive and synergistic partnership.

\emph{Ethical Considerations}.
As ProcessGPT continues to advance, ethical considerations become increasingly important. Future directions for ProcessGPT will prioritize fairness, transparency, and accountability, ensuring that the system operates ethically and respects user privacy and data security.

\emph{Privacy-Preserving Techniques}.
As an ongoing work, we are considering developing privacy-preserving techniques to allow organizations to leverage the benefits of ProcessGPT while ensuring the protection of sensitive or confidential information. This could involve federated learning, differential privacy, and secure multi-party computation techniques.

\emph{Integration with Existing Systems}.
An important line of work will focus on researching techniques to seamlessly integrate ProcessGPT with existing software and systems used in organizations. This would facilitate smooth interoperability and enable ProcessGPT to leverage and enhance the capabilities of existing tools and platforms. This line of work will also involve studying user needs, preferences, and workflows to create a seamless and productive user experience that optimally combines human expertise with ProcessGPT's automation and augmentation capabilities.

Continuous Learning and Improvement.
One promising future direction is to focus on developing sophisticated feedback mechanisms that facilitate user interactions and knowledge exchange. By actively soliciting feedback from users, ProcessGPT can learn from their expertise and incorporate it into its decision-making processes. This feedback loop can involve techniques such as reinforcement learning, where the system learns through trial and error based on the outcomes of its actions, or active learning, which involves selecting informative data points to improve model performance.
Moreover, exploring methods to leverage new data continuously is another crucial avenue of research.

\section*{Acknowledgement}

We acknowledge the Centre for Applied Artificial Intelligence at Macquarie University, Sydney, Australia, for funding this research.

\bibliographystyle{IEEEtran}
\bibliography{bibliography}

\vspace{0cm}
\begin{IEEEbiography}[{\includegraphics[width=1in,height=1.25in,clip,keepaspectratio]{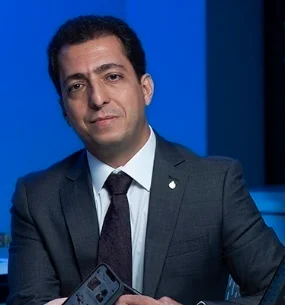}}]{Amin Beheshti}
{\space}is a Full Professor of Data Science, the Director of the Centre for Applied Artificial Intelligence, the head of the Data Science Lab, and the founder of the Big Data Society at Macquarie University, Sydney, Australia. Additionally, he is an Adjunct Professor of Computer Science at UNSW Sydney, Australia. Amin completed his PhD and Postdoc in Computer Science and Engineering at UNSW Sydney and holds a Master's and Bachelor's degree in Computer Science, both with First Class Honours. Amin has significantly contributed to research projects, serving as the R\&D Team Lead and Key Researcher in the 'Case Walls \& Data Curation Foundry' and 'Big Data for Intelligence' projects. These projects were awarded the National Security Impact Award in 2016 and 2017. As a distinguished researcher in Big-Data/Data/Process Analytics, Amin has been invited to serve as a Keynote Speaker, General-Chair, PC-Chair, Organisation-Chair, and program committee member of top international conferences. He is the leading author of several authored books in data, social, and process analytics, co-authored with other high-profile researchers. To date, Amin has secured over \$21 million in research grants for AI-Enabled, Data-Centric, and Intelligence-Led projects.
\end{IEEEbiography}

\vspace{0cm}
\begin{IEEEbiography}[{\includegraphics[width=1in,height=1.25in,clip,keepaspectratio]{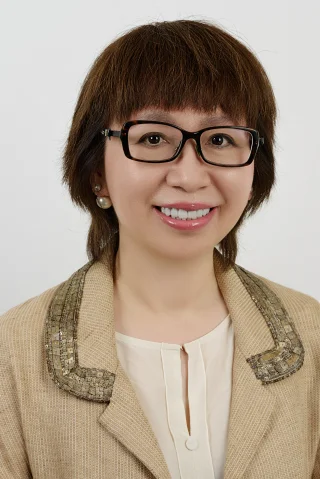}}]{Jian Yang}
is a Full Professor and the Director of Research in the School of Computing, Macquarie University, Sydney, Australia. She received her PhD in Data Integration from The Australian National University in 1995. Before she joined Macquarie University, she worked as an associate professor at Tilburg University, Netherlands (2000-2003), as a senior research scientist at the Division of Mathematical and Information Science, CSIRO, Australia (1998-2000), and as a lecturer at Dept of Computer Science, The Australian Defence Force Academy, University of New South Wales (1993-1998).
Prof Yang's current interests are data-driven process analytics; graph mining; trust, influence, and recommendation in social media; fraud detection; data fusion,  service-oriented computing; smart city; aging and aged care.
Her research has been supported by various sources: ARC Discovery, ARC Linkage, ARC LIEF, CRCs, Data61 CSIRO CRP, and the EU six framework.
\end{IEEEbiography}

\vspace{0cm}
\begin{IEEEbiography}[{\includegraphics[width=1in,height=1.25in,clip,keepaspectratio]{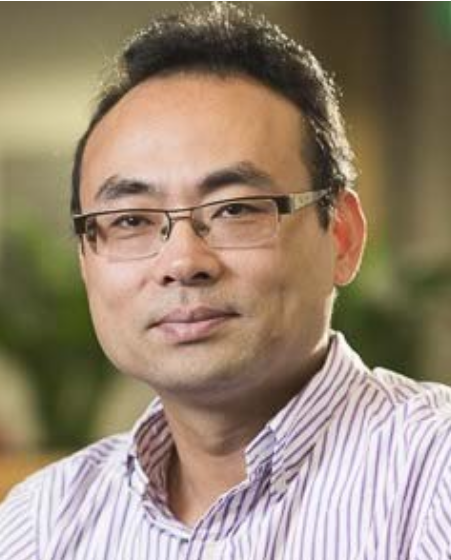}}]{Quan Z. Sheng}{\space}received his Ph.D. degree in computer science from the University of New South Wales, Sydney, NSW, Australia.
He is currently a Full Professor and Head of the School of Computing, at Macquarie University, Sydney. His research interests include big data analytics, service-oriented computing, and the Internet of Things.
Before moving to Macquarie, Michael spent 10 years at the School of Computer Science, the University of Adelaide (UoA), serving in a number of senior leadership roles, including acting Head and Deputy Head of the School of Computer Science. Microsoft Academic ranked Prof. Michael Sheng as one of the Most Impactful Authors in Services Computing (ranked Top 5 All-Time) and in the Web of Things (ranked Top 20 All-Time). He is the recipient of the AMiner Most Influential Scholar Award on IoT (2007-2017), ARC Future Fellowship (2014), Chris Wallace Award for Outstanding Research Contribution (2012), and Microsoft Research Fellowship (2003). Prof Michael Sheng is the Vice Chair of the Executive Committee of the IEEE Technical Community on Services Computing (IEEE TCSVC), the Associate Director (Smart Technologies) of Macquarie's Smart Green Cities Research Centre, and a member of the ACS Technical Advisory Board on IoT.
\end{IEEEbiography}

\vspace{0cm}
\begin{IEEEbiography}[{\includegraphics[width=1in,height=1.25in,clip,keepaspectratio]{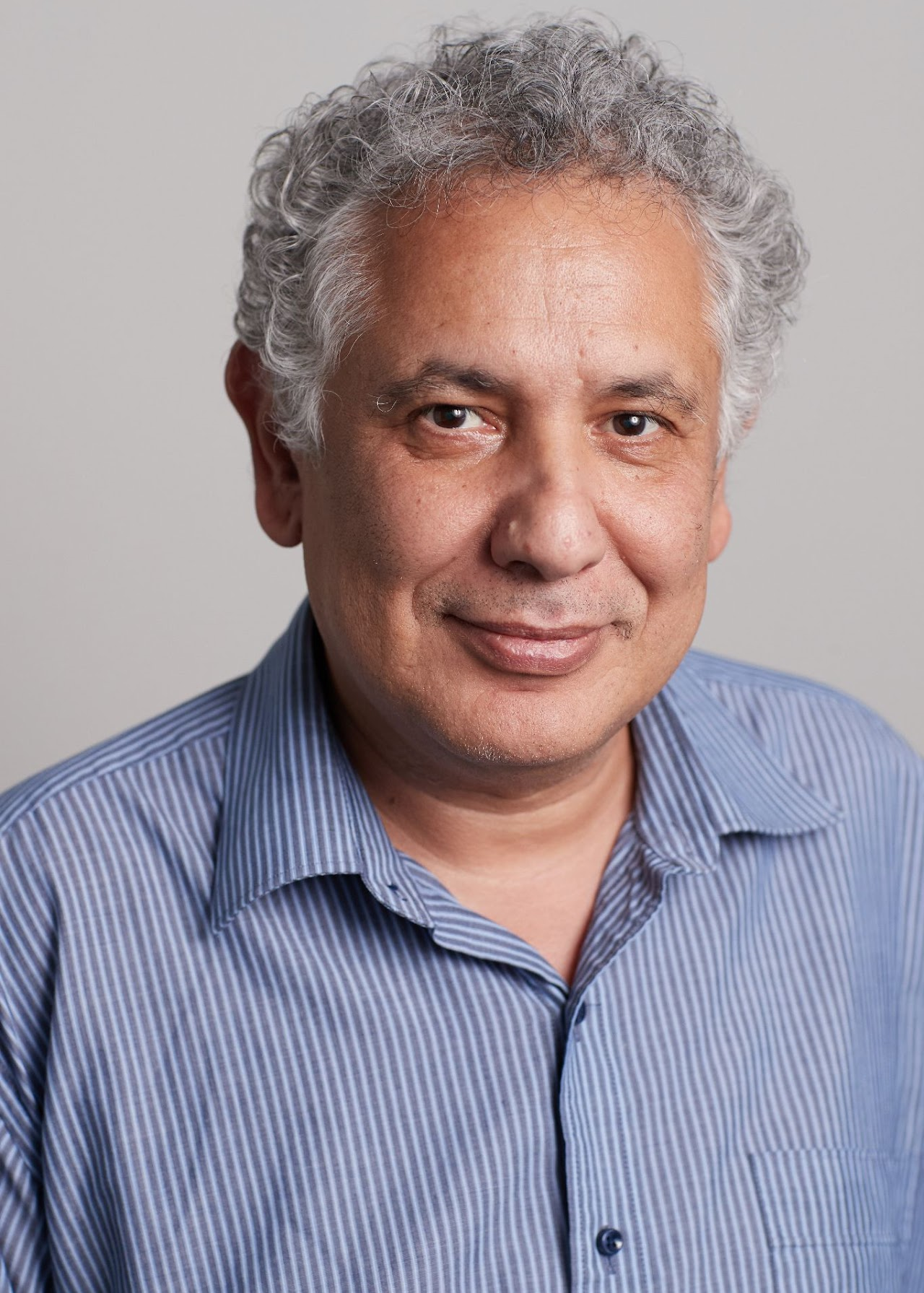}}]{Boualem Benatallah}
{\space} is a Full Professor of Computing at Dublin City University in Ireland, since January 2022.
He has had over 21 years as a Senior Lecturer, an Associate Professor, a Professor, and then a Scientia Professor with UNSW Sydney, Australia.
He is a fellow of the IEEE.
His current research interests revolve around conversational and cognitive services and processes, quality control in crowdsourcing and AI-enabled services. He delves into AI and crowd-based AI training data
curation and quality and employs AI and crowdsourcing services for vulnerability discovery. Furthermore, he explores the intersection of AI and socio-computational services for research and development, cloud services orchestration, and AI-enabled process automation. Boualem serves on the editorial boards of esteemed publications, including the associate editor of ACM Transactions on the Web, IEEE Transactions on Services Computing, and ACM Computing Surveys. He served as associate editor of IEEE Transactions on Cloud Computing. He has played vital roles in various steering committees, notably serving as a member of the ICSOC (International Conference on Service-Oriented Computing) steering committee since 2008 and the BPM (International Conference on Business Process Management) steering committee from 2005 to 2022.
\end{IEEEbiography}

\vspace{0cm}
\begin{IEEEbiography}[{\includegraphics[width=1in,height=1.25in,clip,keepaspectratio]{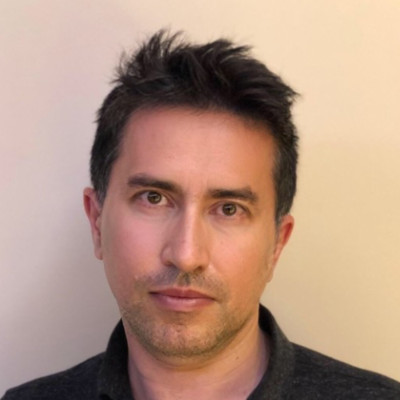}}]{Fabio Casati}
{\space}is  an Honorary Professor at Macquarie Universtiy, Sydney, Australia, and a Full Professor University of Trento in Italy and. Fabio is also the Principal Machine Learning Engineer, Principal Architect, and Leader of AI Trust and Governance Lab at ServiceNow, Palo Alto, CA, USA. Fabio has made significant contributions to various areas, including service-oriented computing, business process management, and software engineering.
His primary scientific interests are in Crowdsourcing, Artificial intelligence, Data Science, Human–computer interaction and Gerontology. His Crowdsourcing study incorporates themes from Quality, Multimedia, Empirical research and Task. His Data science research includes elements of Data-driven, Business processes and Services Computing. His Web service research incorporates themes from Middleware, User interface and Service design.
Fabio has held leadership positions in academia and industry, and his expertise and research have been recognized worldwide.
\end{IEEEbiography}

\vspace{0cm}
\begin{IEEEbiography}[{\includegraphics[width=1in,height=1.25in,clip,keepaspectratio]{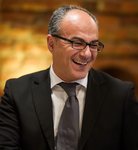}}]{Schahram Dustdar}
{\space}is a Full Professor of Computer Science heading the Research Division of Distributed Systems at TU Wien, Austria. He holds several honorary positions: Francqui Chair Professor at the University of Namur, Belgium (2021-2022), University of California (USC) Los Angeles; Monash University in Melbourne, Shanghai University, Macquarie University in Sydney, University Pompeu Fabra, Barcelona, Spain. From Dec 2016 until Jan 2017 he was a Visiting Professor at the University of Sevilla, Spain and from January until June 2017 he was a Visiting Professor at UC Berkeley, USA.
From 1999 - 2007 he worked as the co-founder and chief scientist of Caramba Labs Software AG in Vienna (acquired by Engineering NetWorld AG), a venture capital co-funded software company focused on software for collaborative processes in teams. Caramba Labs was nominated for several (international and national) awards.
Schahram was the founding co-Editor-in-Chief of ACM Transactions on Internet of Things (ACM TIoT) and is the Editor-in-Chief of Computing (Springer). He is an Associate Editor of IEEE Transactions on Services Computing, IEEE Transactions on Cloud Computing, ACM Computing Surveys, ACM Transactions on the Web, and ACM Transactions on Internet Technology, as well as on the editorial board of IEEE Internet Computing and IEEE Computer. Dustdar is the recipient of multiple awards, including: TCI Distinguished Service Award (2021), IEEE TCSVC Outstanding Leadership Award (2018), IEEE TCSC Award for Excellence in Scalable Computing (2019), ACM Distinguished Scientist (2009), and IBM Faculty Award (2012).
\end{IEEEbiography}

\vspace{0cm}
\begin{IEEEbiography}[{\includegraphics[width=1in,height=1.25in,clip,keepaspectratio]{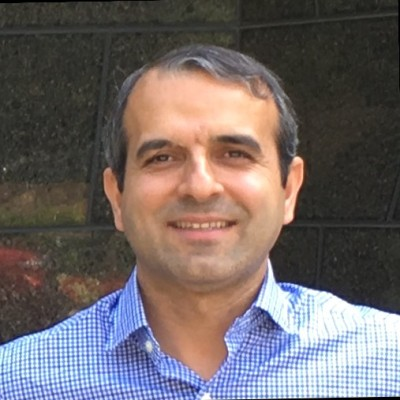}}]{Hamid Reza Motahari Nezhad}
{\space}is an Honorary Professor at the School of Computing, Macquarie University, Sydney, Australia.
Hamid is the Founder, Chief Scientist and CEO at UpBrains AI, Inc., Washington, USA; with a focus on AI for business processes, document intelligence, and conversational AI. Before UpBrains AI, Hamid served as Head of AI Science at EY AI Lab, Palo Alto, in EY Global Technology Innovation where he led AI science team, and intellectual property and AI-based innovation programs, and prior to that he was Research Lead for Cognitive AI services at IBM Research where he led successful R\&D-based product delivery to IBM software, Watson and services business units.
Hamid has extensive scholarly publications, dozens of US patents, and experience in leading and managing applied AI science and engineering teams, strategic leadership and technical program management, thought leadership and delivery of innovative AI products and solutions within corporate R\&D bodies. During his tenure with IBM Research, Hamid was a member of the IBM Academy of Technology, which is a selected community of technology leaders across IBM.
\end{IEEEbiography}

\vspace{0cm}
\begin{IEEEbiography}[{\includegraphics[width=1in,height=1.25in,clip,keepaspectratio]{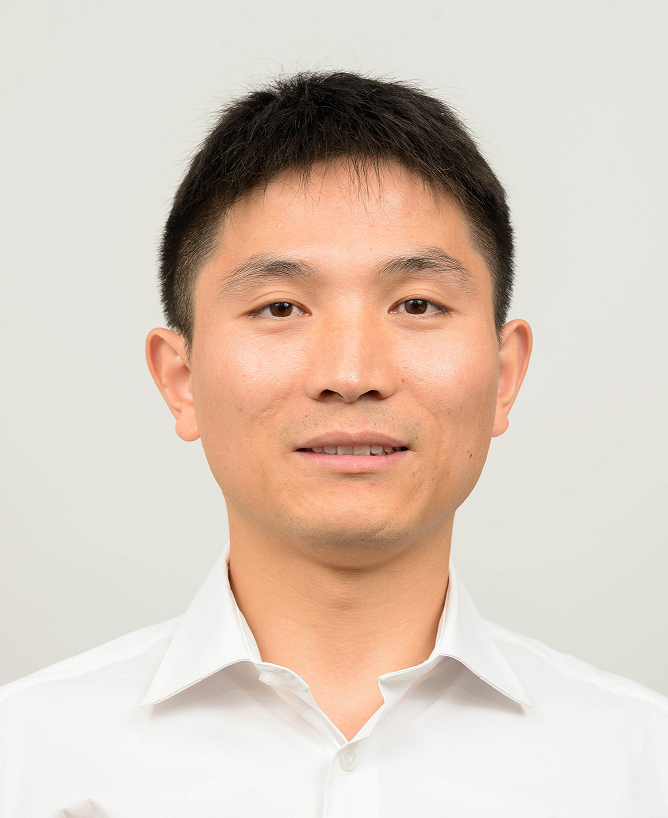}}]{Xuyun Zhang}
{\space}is a distinguished ARC Discovery Early Career Research Awardee (DECRA) and currently holds the position of Senior Lecturer in the School of Computing at Macquarie University in Australia. Previously, he served as a Lecturer at The University of Auckland from 2016 to 2019 and as a Postdoctoral Researcher at NICTA (National ICT Australia, now Data61, CSIRO) from 2014 to 2016. Dr. Zhang obtained his Ph.D. degree in Computer Science and Technology from the University of Technology Sydney (UTS) in Australia in 2014, following his Master's and Bachelor's degrees in the same field from Nanjing University, China in 2011 and 2008 respectively. Throughout his academic journey, Xuyun has demonstrated research excellence and leadership potential, earning him the prestigious 2019 Research Excellence Awards during his tenure at the University of Auckland. His research interests encompass various domains, including scalable and secure machine learning, big data privacy and cybersecurity, big data mining and analytics, cloud/edge/service computing, and IoT.
\end{IEEEbiography}

\vspace{0cm}
\begin{IEEEbiography}[{\includegraphics[width=1in,height=1.25in,clip,keepaspectratio]{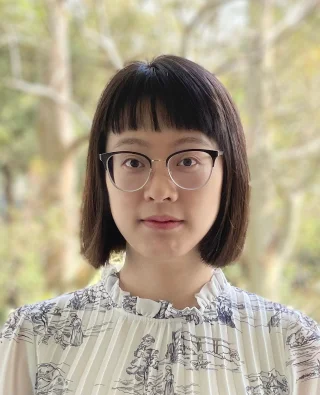}}]{Shan (Emma) Xue}
{\space} is a Lecturer in Computing at Macquarie University, Sydney, Australia. Dr. Xue holds a Ph.D. in Software Engineering from the University of Technology Sydney (2019) and a Ph.D. in Information Management and Information System from Shanghai University (2018). Emma worked as a Lecturer at the University of Wollongong in 2022 and conducted postdoctoral research at Macquarie University with joint support from CSIRO Data61 from 2019 to 2022. Her research interests lie in the realm of deep learning, adaptive artificial intelligence, data mining, and knowledge discovery in complex cyber environments. With a focus on graph analytics, graph representation, deep learning, and business/web intelligence, her research outputs have been published in prestigious international conferences such as NeurIPS, IJCAI, WSDM, ICDM, and esteemed journals including Pattern Recognition, The VLDB Journal,  and IEEE Transactions on Neural Networks and Learning Systems (TNNLS). Emma actively contributed to the research community by serving as a general sessional chair at IJCNN and holding positions as SPC/PC member in top international conferences such as AAAI, IJCAI, WWW, WSDM, and KDD.
\end{IEEEbiography}

\end{document}